\documentclass[11pt]{article}

\usepackage[preprint]{acl}

\usepackage{times}
\usepackage{latexsym}

\usepackage[T1]{fontenc}
\usepackage{times}
\usepackage{latexsym}
\usepackage{booktabs}
\usepackage{multirow}
\usepackage{amssymb}
\usepackage{amsmath}

\usepackage[utf8]{inputenc}

\usepackage{microtype}

\usepackage{inconsolata}

\usepackage{graphicx}

\title{From Static Analysis to Audience Dissemination: A Training-Free Multimodal Controversy Detection Multi-Agent Framework}

\author{
  Zihan Ding\textsuperscript{1}
  \and
  Ziyuan Yang\textsuperscript{2}
  \and
  Yi Zhang\textsuperscript{1}\thanks{*Corresponding author: \href{mailto:yzhang@scu.edu.cn}{yzhang@scu.edu.cn}}
  \\[6pt] % 作者行和单位之间增加间距，更美观
  \textsuperscript{1}Sichuan University, Chengdu, Sichuan, China \\
  \texttt{2022141230168@stu.scu.edu.cn}, \texttt{yzhang@scu.edu.cn}
  \\[4pt]
  \textsuperscript{2}Nanyang Technological University, Singapore \\
  \texttt{cziyuanyang@gmail.com}
}

\begin{document}
\maketitle
\begin{abstract}
Multimodal controversy detection~(MCD) detects controversies in videos and their associated comments to support risk management on social video platforms.
Prior work treats MCD as a static representation learning problem, directly extracting features from videos and their associated comments. However, these methods fail to capture diverse evaluations from different audiences. Inspired by real-world audience dissemination, we propose a training-free multi-agent framework \textit{AuDisAgent} that models MCD as a dynamic propagation process.
Our framework explicitly models audience dissemination through a structured multi-agent system. First, three \textit{Screening Agents}, including the \textit{Video}, \textit{Comment}, and \textit{Interaction Agents}, perform initial assessments from visual, textual, and cross-modal perspectives, respectively. For samples where consensus cannot be reached, the \textit{Viewing Panel Agent} is activated to simulate a post-screening discussion among audiences with diverse backgrounds and stances. This mechanism models how different audience groups interpret and react to the same content, exposing latent controversy contents that may emerge during dissemination.
Finally, the \textit{Arbitration Agent} makes the final judgment based on the preceding reasoning chain.
In addition, to address the “cold-start” scenario where newly released videos lack comments, we implement a \textit{Comment Bootstrapping Strategy} that transfers similar historical public comments as the initial comment. Extensive experiments on the public dataset show that our framework significantly outperforms existing state-of-the-art~(SOTA) methods in both rich- and limited-comment scenarios.
\end{abstract}

\section{Introduction}

The rapid rise of short-video platforms, such as TikTok and YouTube Shorts, has fundamentally reshaped the information dissemination and social interaction \cite{Wang_Gan_Wei_Wu_Meng_Nie_2022}. While these platforms facilitate valuable content dissemination, they also serve as breeding grounds for online controversies where polarized views can rapidly escalate into reputational damage or social conflicts \cite{Wang_Li_Gan_Zhang_Lv_Nie_2023,hessel-lee-2019-somethings}. Therefore, Multimodal Controversy Detection (MCD), which aims to assess whether a video and its associated comments are controversial, has become critical for content governance and risk management on social platforms~\cite{Cook_Jull_Moore_2014}. 

Existing MCD methods treat this task as a static representation learning problem, directly extracting features from videos and associated comments to detect controversy\cite{Sanh_Webson_Raffel_Bach_Sutawika_Alyafeai_Chaffin_Stiegler_Scao_Raja_et_al._2021}. 
For example, MVCD \cite{Xu_Chen_Zhao_Gao_Gan_2024} and TPC-GCN \cite{Zhong_Cao_Sheng_Guo_Wang_2020} use graph neural networks to statically encode the features for controversy detection. Moreover, recent works utilize Large Language Models (LLMs) \cite{Ma_Wang_Xing_Zhao_Zhang_2024,Tahmasebi_Muller_Budack_Ewerth_2024} to align visual and textual features from videos and comments for MCD.
Despite recent progress, mainstream methods treat controversy as a \textit{static} content attribute~\cite{tsai-etal-2019-multimodal,Garimella_De_Francisci_Morales_Gionis_Mathioudakis_2016}. However, in practice, controversy is not a static property but a dynamic social phenomenon emerging from the interactions and opinion conflicts among audiences. Hence, static learning-based models often remain limited to capture the underlying controversy.

To address these issues, it is essential to move beyond “\textit{static feature modeling}” toward “\textit{dynamic simulation of audience dissemination}”. Only by simulating the dynamic process can we uncover how controversies are shaped by complex interactions among diverse audiences. Motivated by this, we propose a training-free and interpretable multi-agent framework \textit{AuDisAgent} for MCD that simulates audience dissemination to reveal how controversies emerge from complex interactions among diverse audiences.

\begin{figure}[!t]
 \small
  \centering
  \includegraphics[width=\linewidth]{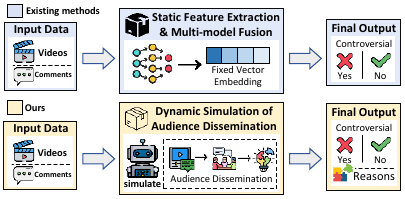}  
  \vspace{-5pt}
  \caption{Pipeline comparison between existing MCD methods and our proposed approach.}
  \vspace{-10pt}
  \label{fig:concept}
\end{figure}

Specifically, \textit{AuDisAgent} explicitly models the audience dissemination process through a structured multi-agent system. First, three \textit{Screening Agents}, including the \textit{Video}, \textit{Comment}, and \textit{Interaction Agents}, perform initial assessments from visual, textual, and cross-modal perspectives, respectively. For ambiguous samples where consensus cannot be reached, the \textit{Viewing Panel Agent} will be activated to simulate a post-screening discussion among audiences with diverse backgrounds.
Instead of only producing a decision, each agent also provides the rationale behind it, and these justifications collectively form the reasoning chain.
Finally, the \textit{Arbitration Agent} integrates the preceding reasoning chain to make the final controversy judgment and provides an interpretable decision basis. To clarify our method and its differences from previous work, we illustrate the concept with an example in Fig.~\ref{fig:concept}.

Furthermore, to address the “\textit{cold-start}” scenario where newly released videos lack comments, we introduce a \textit{Comment Bootstrapping Strategy}. 
By selecting the comments from the semantically similar videos as initial comments, this strategy constructs a preliminary corpus to guide agents in simulating potential future public opinion~\cite{Xu_Chen_Zhao_Gao_Gan_2024,ijcai2025p1107}. Extensive experiments on the public dataset demonstrate that AuDisAgent significantly outperforms existing baseline methods in both rich- and limited-comment scenarios. In summary, our main contributions are as follows:
\begin{itemize}
    % \item We identify the limitations of existing MCD methods, which treat the task as a static representation learning problem, and propose a shift toward modeling controversy as a dynamic audience dissemination process.
    \item We introduce a training-free multi-agent framework AuDisAgent to reformulate the MCD task from static feature modeling to dynamic simulation of audience dissemination.
    % leverages Screening Agents, a Viewing Panel Agent, and an Arbitration Agent to simulate post-screening discussions and mine latent controversial contents.
    \item We propose a Comment Bootstrapping Strategy to resolve the cold-start scenario that retrieves historical public comments from semantically similar videos.
    \item Extensive experiments on the public dataset demonstrate that our method outperforms state-of-the-art~(SOTA) methods.
    % , with the additional advantage of providing clear, interpretable reasoning paths for risk management.
\end{itemize}

\begin{figure*}[!t]
 \small
  \centering
  \includegraphics[width=\linewidth]{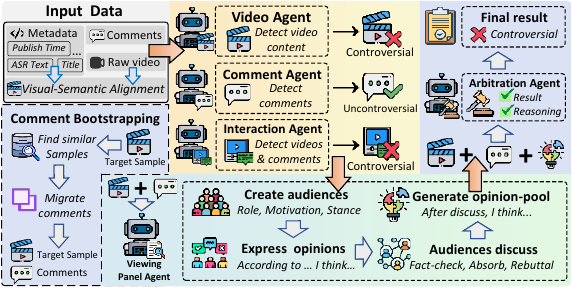}  
  \caption{Overview of the proposed AuDisAgent.}
  \label{fig:overview}
  \vspace{-5pt}
\end{figure*}

\section{Related Work}

\subsection{Controversy Detection}
Driven by the proliferation of user-generated content, MCD has become crucial for risk management and content moderation \cite{Zhong_Cao_Sheng_Guo_Wang_2020,hessel-lee-2019-somethings,Linmans_van_de_Velde_Kanoulas_2018}. Early studies rely on manually constructed dictionaries or rules to identify predefined topics \cite{Garimella_Morales_Gionis_Mathioudakis_2018}, which struggle to capture the evolving and context-dependent nature of controversies. Subsequent research addressed this by incorporating multi-dimensional features like semantics and social background \cite{Linmans_van_de_Velde_Kanoulas_2018,Zhong_Cao_Sheng_Guo_Wang_2020}, while application scenarios expanded to complex, dynamic multimodal environments such as social media \cite{Linmans_van_de_Velde_Kanoulas_2018,ijcai2025p1107}. 
Methodologically, approaches evolved from statistical rules \cite{Popescu_Pennacchiotti_2010,Hamad_Skowron_Schedl_2018} to graph-based analysis \cite{Mendoza_Parra_Soto_2020} and, recently, pre-trained neural networks for multimodal detection \cite{Xu_Chen_Zhao_Gao_Gan_2024}. However, existing models focus on binary classification and lack interpretability.
% To address this, we introduce generative agents to provide transparent, interpretable reasoning for controversies.

\subsection{Multi-Agent Framework}
LLMs have gained extensive attention as autonomous agents due to their strong planning and reasoning capabilities \cite{Yao_2022,Shen_2023,Mu_2023,Zhao_2024,Sun_2023}. They effectively handle complex tasks under weakly supervised conditions, reducing reliance on labeled data. Building on this, multi-agent frameworks \cite{Qian_2024,Tao_2024,Hong_2024_CVPR,Ma_2024_CVPR} enable interactive collaboration, proving more effective than single-agent systems for collective problem-solving. 
However, their application in MCD remains scarce. For instance, AgentMCD \cite{Xu_Chen_Zhao_Gao_Gan_2024} applies multi-agent technology to multimodal detection, but it lacks deep interactive reasoning and fails to leverage simulated social interactions to uncover core controversies.
\section{Methodology}

\subsection{Overview}
Given a short video post, MCD methods aim to determine whether the content poses a potential risk of public opinion conflict. Formally, a sample is represented as $\mathcal{S}=\{v,\mathcal{T}_{meta},C\}$, where $v$ denotes the video, $\mathcal{T}_{meta}$ contains textual metadata (e.g., title, keywords, publisher information), and $C=\{c_{1},c_{2},...,c_{n}\}$ denotes the comment set. We aim to design an interpretable framework that provides both the reasoning process and the final prediction.
% :\mathcal{D}\rightarrow\{y,\mathcal{E}\}$, where $y\in\{0,1\}$ is a binary controversy prediction and $\mathcal{E}$ is a natural language rationale.$

In this paper, we propose a novel training-free multi-agent framework \textit{AuDisAgent} that reformulates the MCD task from static feature modeling to dynamic simulation of audience dissemination. The overview of AuDisAgent is shown in Fig.~\ref{fig:overview}. Specifically, three \textit{Screening Agents}~(Video, Comment, and Interaction) first conduct assessments and generate distinct reasoning chains. For ambiguous cases, the \textit{Viewing Panel Agent} simulates discussions among diverse audiences to uncover latent controversies. Then, the \textit{Arbitration Agent} integrates these rationales to produce the reasoning process and the final prediction. Furthermore, to address the “cold-start” videos, we introduce a \textit{Comment Bootstrapping Strategy} that leverages relevant historical data to simulate the potential audience dissemination.

\subsection{Visual-Semantic Alignment}
To enable semantic reasoning , we first align visual information with textual representations.
We employ mPLUG-Video \cite{xu2023youkumplug} as our visual encoder $\mathcal{F}$, which is pretrained on large-scale corpus of paired video–text data. We construct a comprehensive prompt $\mathcal{P}_{desc}$ to generate a refined video description $\mathcal{T}_v$, which can be formulated as:
\begin{equation}
\mathcal{T}_v = \mathcal{F}(v, \mathcal{T}_{meta}, \mathcal{P}_{desc}).
\end{equation}

% $\mathcal{T}_v$ denotes the textual representation of the video.
The metadata is introduced to avoid relying solely on visual content, which may yield coarse or underspecified descriptions.
Hence, $\mathcal{T}_v$ integrates the visual representation of the video with its textual metadata.
% For example, when a reporter expresses personal opinions on camera, the generated description might merely be "a person speaking to the camera". This kind of description lacks the semantic information to capture the discussed topics or articulated viewpoints. Incorporating the metadata can effectively resolve this issue and yield descriptions with richer details.

\subsection{Screening Agents}
To model diverse first impressions and capture modality-specific controversy cues, we deploy three specialized \textbf{Screening Agents} to perform preliminary judgments from different modal perspectives:
1) \textbf{Video Agent} $\mathcal{A}_{v}$ operates exclusively on the video to assess the intrinsic semantic information indicating potential controversy.
2) \textbf{Comment Agent} $\mathcal{A}_{c}$ processes the comments to identify strong disagreements or conflicting viewpoints among the past audiences.
3) \textbf{Interaction Agent} $\mathcal{A}_{i}$ analyzes the joint context of the video and the comments to detect cross-modal controversy between the video and the audience's reaction.

Each agent independently evaluates its input and predicts whether the content is controversial.
Then, AuDisAgent utilizes a consistency gating mechanism to detect controversial content: if all three screening agents provide the same decision, the reasoning process concludes and gives the consensus decision. However, if the agents do not reach a consensus, the sample is treated as an ambiguous case and moves forward to the next agent to model the dynamic dissemination process for further analysis.

\subsection{Viewing Panel Agent}
For ambiguous samples, static content alone is often insufficient to determine whether a video is controversial. In such cases, controversy may emerge from interactions among opposing viewpoints during audience dissemination. Hence, we design the \textbf{Viewing Panel Agent} $\mathcal{A}_{p}$ to simulate the discussions among audiences with diverse backgrounds.

Specifically, we first extract diverse audience personas $\mathcal{P}$ from the video representation $\mathcal{T}_v$ and the filtered comment set $\mathcal{C}'$. To reduce noise, we only retain the most liked comments for each video; the top 30 liked comments are empirically selected to balance performance and efficiency. Each persona is characterized by three attributes: \textit{social role}, \textit{motivation}, and \textit{core stance}.
Then, each persona expresses the initial opinion $\mathcal{O}$ and engages in discussion to simulate audience dissemination. During the discussion, personas may perform three operations: 1) \textit{Fact-checking:} Correcting hallucinations using video evidence; 2) \textit{Absorption:} Adopting reasonable logic from others; 3) \textit{Rebuttal:} Refuting unreasonable viewpoints to maintain a stance.
After the discussion, each persona updates its opinion to $\mathcal{O}'$, which is added to the opinion pool $\mathcal{H}'$.

\subsection{Arbitration Agent}
Once the simulated audience dissemination is completed, the \textbf{Arbitration Agent} $\mathcal{A}_{j}$ produces the final decision. Instead of relying solely on raw inputs, it integrates $\mathcal{T}_v$, $\mathcal{C}'$, and the final evolved opinion pool $\mathcal{H}'$. 
Based on these, the Arbitration Agent produces the controversy score and the corresponding explanation. This process can be formulated as follows:
\begin{equation}
S, \mathcal{E} = \mathcal{A}_{j}(\mathcal{T}_v, \mathcal{C}', \mathcal{H}'),
\end{equation}
where $S$ denotes the predicted controversy score, and $\mathcal{E}$ is the interpretable decision reasoning chain.

Then, the final controversy prediction is obtained as $\hat{y} = \mathbb{I}(S \geq \tau)$, where $\tau$ is the decision threshold, which is set to the median of the scoring range~(0–25) specified in the prompt.

\subsection{Comment Bootstrapping Strategy}

The above process assumes the availability of an initial comment set to support audience dissemination simulation. However, in real-world scenarios, newly published videos may face a cold-start situation where few or no comments are available~(i.e., $\mathcal{C}_{q} \approx \emptyset$). In such cases, the Viewing Panel Agent lacks sufficient audience feedback, making it difficult to simulate opinion evolution and dissemination dynamics.
To address this, we design a \textit{Comment Bootstrapping Strategy}. Specifically, for a target sample $\mathcal{S}_{q}$ without comments, a pretrained model bge-large-zh-1.5~\cite{xiao2024cpack}, denoted as $\mathcal{H}(\cdot)$), is utilized to encode the “title” within the corresponding $\mathcal{T}_{meta}$ to get $\mathbf{E}_{q}$. The same encoding process is applied to all samples in the reference database $\mathcal{D}_{ref}$ as follows:
\begin{equation}
    \mathbf{E}_{ref}^{i} = \mathcal{H}(\mathcal{S}_{ref}^{i}),\;
\mathcal{S}_{ref}^{i} \in \mathcal{D}_{ref},
\end{equation}
where $\mathbf{E}_{ref}^{i}$ denotes the feature of $\mathcal{S}_{ref}^{i}$.

Then, we compute the cosine similarity between the query embedding $\mathbf{E}_{q}$ and each reference embedding in the database to retrieve the top-3 most similar samples. For each sample, the 10 most-liked comments are selected to construct a proxy comment set $C_{proxy}$ for newly posted videos without comments. This proxy set serves as prior feedback to guide our framework in simulating potential future audience reactions.

\section{Experiments}
\begin{table*}[t!]
\small

% \vspace{-5pt}
\centering
% \scalebox{1.2}{
\begin{tabular}{lcccccccc}
\toprule
& \multicolumn{4}{c}{\textbf{with rich comments}} & \multicolumn{4}{c}{\textbf{with limited comments}} \\
\cmidrule(lr){2-5} \cmidrule(lr){6-9}
\textbf{Method} & F1 & Rec. & Prec. & Acc. & F1 & Rec. & Prec. & Acc. \\
\midrule
Standard Prompting & 67.91 & 67.93 & 67.98 & 67.93 & 64.39 & 64.40 & 64.42 & 64.40 \\
Zero-shot CoT \cite{Kojima_Gu_Reid_Matsuo_Iwasawa_2022} & 69.15 & 69.17 & 69.22 & 69.17 & 64.02 & 64.05 & 64.09 & 64.05 \\
Plan-and-Solve \cite{Wang_Xu_Lan_Hu_Lan_Lee_Lim_2023} & 68.11 & 68.11 & 68.12 & 68.11 & 62.87 & 63.25 & 63.81 & 63.25 \\
Self-Consistency \cite{Wang_Li_Zhao_2023} & 65.76 & 65.83 & 65.95 & 65.83 & 61.75 & 61.75 & 61.75 & 61.78 \\
Self-Reflect  \cite{Shinn_Cassano_Berman_Gopinath_Narasimhan_Yao_2023} & 69.13 & 69.26 & 69.58 & 69.26 & 63.96 & 63.96 & 63.96 & 63.96 \\
Self-Refine \cite{Madaan_Tandon_Gupta_Hallinan_Gao_Wiegreffe_Alon_Dziri_Prabhumoye_Yang_et_al._2023} & 62.08 & 63.96 & 67.41 & 63.96 & 61.78 & 63.25 & 65.66 & 63.25 \\
Cumulative Reasoning (CR) \cite{Zhang_Yang_Yuan_Yao_2023} & 50.16 & 57.42 & 67.78 & 57.42 & 58.22 & 59.72 & 61.34 & 59.72 \\
RECITE \cite{Jiang_Xu_Gao_Sun_Liu_Dwivedi-Yu_Yang_Callan_Neubig_2023} & 61.89 & 63.03 & 64.82 & 63.04 & 53.14 & 57.42 & 61.69 & 57.42 \\
Tree of Thoughts (ToT) \cite{Yao_Yu_Zhao_Shafran_Griffiths_Cao_Narasimhan_2023} & 68.41 & 68.46 & 68.58 & 68.46 & 60.81 & 61.61 & 62.66 & 61.63 \\
AutoAgents \cite{Chen_Dong_Shu_Zhang_Sesay_Karlsson_Fu_Shi_2023} & 64.17 & 65.20 & 65.93 & 64.39 & 60.14 & 60.28 & 60.42 & 60.27 \\
Multi-Agent (Debate) \cite{Liu_Du_Li_Tenenbaum_Torralba_2023} & 63.44 & 64.27 & 65.68 & 64.25 & 60.46 & 60.75 & 61.07 & 60.74 \\
MAD \cite{Liang_He_Jiao_Wang_Wang_Wang_Yang_Shi_Tu_2024} & 61.84 & 61.84 & 61.85 & 61.85 & 62.58 & 62.63 & 62.71 & 62.63 \\
AgentMCD \cite{ijcai2025p1107} & 69.98 & 70.14 & 70.60 & 70.14 & 66.29 & 66.45 & 66.78 & 66.46 \\
\midrule
 AuDisAgent (Ours) & \textbf{71.64} & \textbf{72.08} & \textbf{71.20} & \textbf{71.47} & \textbf{68.29} & \textbf{69.75} & \textbf{66.89} & \textbf{67.56} \\
\bottomrule
\end{tabular}
% }
% \vspace{-5pt}

\caption{Performance comparison among different methods on MMCD.}
\label{tab:mmcd_performance}
\end{table*}

\begin{table*}[t]
\centering
\small

% 降低缩放比例，避免宽度溢出；也可根据需求调整为1.0/1.1
% \scalebox{1.2}{
% \vspace{-5pt}
\begin{tabular}{l cccc cccccccc}
\toprule
% 修复：\multirow{2}{*} 表示跨2行，*表示自动适配宽度
\multirow{2}{*}{\textbf{Method}} 
& $\mathcal{A}_v$ & $\mathcal{A}_{i}$ & $\mathcal{A}_c$ & $\mathcal{A}_p$
& \multicolumn{4}{c}{\textbf{with rich comments}} 
& \multicolumn{4}{c}{\textbf{with limited comments}} \\
\cmidrule(lr){2-5} \cmidrule(lr){6-9} \cmidrule(lr){10-13}
& & & & & F1 & Rec. & Prec. & Acc. & F1 & Rec. & Prec. & Acc. \\
\midrule
$\mathcal{A}_v$-only & \checkmark & & & & 63.27 & 61.48 & 65.17 & 64.31 & 63.10 & 61.48 & 64.80 & 64.05 \\
$\mathcal{A}_{i}$-only  & & \checkmark & & & 65.36 & 64.66 & 66.06 & 65.72 & 60.03 & 63.43 & 56.98 & 57.77 \\
$\mathcal{A}_c$-only  & & & \checkmark & & 68.64 & 67.67 & 69.64 & 69.08 & 63.38 & 68.20 & 59.20 & 60.60 \\
$\mathcal{A}_p$-only &  & &  & \checkmark & 70.03 & 69.96 & 70.09 & 70.05 & 64.69 & 65.37 & 64.01 & 64.31 \\
\midrule
w/o $\mathcal{A}_v$ & &\checkmark & \checkmark & \checkmark &  70.01 & 69.08 & 70.96 & 70.41 & 63.14 & 65.37 & 61.06 & 61.84 \\
w/o $\mathcal{A}_{i}$ & \checkmark &  & \checkmark & \checkmark & 70.63 & 70.32 & 70.94 & 70.76 & 66.37 & 66.78 & 65.97 & 66.17 \\
w/o $\mathcal{A}_c$ & \checkmark & \checkmark &  & \checkmark & 70.16 & 71.02 & 69.31 & 69.79 & 65.20 & 65.37 & 65.03 & 65.11 \\
w/o $\mathcal{A}_p$   & \checkmark & \checkmark & \checkmark &  & 68.12 & 66.25 & 70.09 & 68.99 & 63.55 & 67.14 & 60.32 & 61.48 \\
\midrule
\textbf{ AuDisAgent} & \textbf{\checkmark} & \textbf{\checkmark} & \textbf{\checkmark} & \textbf{\checkmark} & \textbf{71.64} & \textbf{72.08} & \textbf{71.20} & \textbf{71.47} & \textbf{68.29} & \textbf{69.75} & \textbf{66.89} & \textbf{67.56} \\
\bottomrule
\end{tabular}
% }
% \vspace{-5pt}

\caption{Experimental results of the ablation study.}
\label{tab:mmcd_ablation_study}
\end{table*}

\subsection{Experiment Setup}
\textbf{Datasets}: We evaluate the performance on the public MMCD dataset \cite{Xu_Chen_Zhao_Gao_Gan_2024}. It is a large-scale MCD Dataset consisting of over 10,000 Chinese videos, each accompanied by extensive social context information.

\noindent \textbf{Baselines}: We compare AuDisAgent against 13 advanced baselines, encompassing various reasoning and multi-agent paradigms. These include Standard Prompting \cite{glm2024chatglm}, Zero-shot Chain of Thought \cite{Kojima_Gu_Reid_Matsuo_Iwasawa_2022}, Plan-and-Solve\cite{Wang_Xu_Lan_Hu_Lan_Lee_Lim_2023}, Self-Consistency \cite{Wang_Li_Zhao_2023}, Self-Reflect \cite{Shinn_Cassano_Berman_Gopinath_Narasimhan_Yao_2023}, SelfRefine \cite{Madaan_Tandon_Gupta_Hallinan_Gao_Wiegreffe_Alon_Dziri_Prabhumoye_Yang_et_al._2023}, Tree of Thoughts  \cite{Yao_Yu_Zhao_Shafran_Griffiths_Cao_Narasimhan_2023}, Cumulative Reasoning \cite{Zhang_Yang_Yuan_Yao_2023}, RECITE \cite{Jiang_Xu_Gao_Sun_Liu_Dwivedi-Yu_Yang_Callan_Neubig_2023}, AutoAgents \cite{Chen_Dong_Shu_Zhang_Sesay_Karlsson_Fu_Shi_2023}, Multi-Agent \cite{Liu_Du_Li_Tenenbaum_Torralba_2023}, MAD \cite{Liang_He_Jiao_Wang_Wang_Wang_Yang_Shi_Tu_2024}, and AgentMCD \cite{ijcai2025p1107}.

\noindent \textbf{Implementation Details}: Our proposed framework is implemented using a Python script, utilizing GLM4-9B \cite{glm2024chatglm} as the backbone LLM, which supports long-text reasoning up to 128K tokens. In the process of similar video retrieval, we employ the bge-large-zh model \cite{xiao2024cpack} for encoding. All experiments are conducted in one-shot or few-shot settings without additional training or fine-tuning steps. For all experimental cases, the binary classification decision threshold is set to the median of the model's output controversy score range.

\noindent \textbf{Metrics}: Precision, recall, accuracy, and F1-score are adopted as evaluation metrics, where higher values indicate better performance.

\subsection{Main Result}
We compare AuDisAgent with several existing methods on the public MMCD dataset under both rich- and limited-comment scenarios. 
As shown in Table \ref{tab:mmcd_performance}, AuDisAgent consistently outperforms all baseline models in both scenarios. Specifically, compared with the second-best method AgentMCD, AuDisAgent improves accuracy by 1.33\% and F1-score by 1.68\% in the rich-comment scenario. In the limited-comment scenario, it further achieves gains of 1.10\% in accuracy and 2.00\% in F1-score. These results demonstrate the effectiveness of our framework, which stems from reframing MCD from static feature modeling to dynamic simulation of audience dissemination. 

\begin{table}[!t]

\centering
% \vspace{-5pt}
\resizebox{\columnwidth}{!}{
\begin{tabular}{lcccc}  % 总计5列：l + cccc
\toprule
\textbf{Models} & \multicolumn{4}{c}{\textbf{with rich comments}} \\
\cmidrule(lr){2-5}     % 2-5列的横线（匹配4列跨列）
 & F1 & Rec. & Prec. & Acc. \\
\midrule  % 替换\hline，和booktabs风格统一
% 关键修改1：将7改为5，匹配表格总列数；保持c（居中）
\multicolumn{5}{c}{GLM4-9B \cite{glm2024chatglm}} \\
\midrule  % 替换\hline
Baseline & 67.91 & 67.93 & 67.98 & 67.93 \\
AgentMCD  & 69.98 (+2.07) & 70.14 (+2.21) & 70.60 (+2.62) & 70.14 (+2.21) \\
 AuDisAgent  & \textbf{71.64 (+3.73)} & \textbf{72.08 (+4.15)} & \textbf{71.20 (+3.22)} & \textbf{71.47 (+3.54)} \\
\midrule
\multicolumn{5}{c}{Qwen2.5-7B \cite{qwen2025qwen25technicalreport}} \\
\midrule  % 替换\hline
Baseline & 65.76 & 68.37 & 63.34 & 64.40 \\
AgentMCD  & 64.50 (-1.26)  & 63.25 (-5.12) & 65.81 (+2.47) & 65.19 (+0.79)  \\
 AuDisAgent  & \textbf{67.78 (+2.02)} & \textbf{69.11 (+0.74)} & \textbf{66.50 (+3.14)} & \textbf{66.58 (+2.18)}\\
\midrule
\multicolumn{5}{c}{DeepSeek-R1-Distill-Llama-8B\cite{deepseekai2025deepseekr1incentivizingreasoningcapability,llama32024}} \\
\midrule  % 替换\hline
Baseline & 63.73 & 66.43 & 61.24 & 62.19 \\
AgentMCD  & 64.37 (+0.64) & 63.07 (-3.36) & 66.48 (+5.24) & 65.64 (+3.45)   \\
 AuDisAgent  & \textbf{67.66 (+3.93)} & \textbf{68.32 (+1.89)} & \textbf{67.01 (+5.77)} & \textbf{67.40 (+5.21)} \\
\midrule
\multicolumn{5}{c}{gpt-4o \cite{openai2024gpt4osystem}} \\
\midrule  % 替换\hline
Baseline & 67.75  & 66.43  & 69.12 & 68.37  \\
AgentMCD  & 69.20 (+1.45) & 67.49 (+1.06) & 71.00 (+1.88) & 69.96 (+1.59) \\
 AuDisAgent  & \textbf{70.84 (+3.09)} & \textbf{69.96 (+3.53)} & \textbf{71.74 (+2.62)} & \textbf{71.20 (+2.83)} \\
\bottomrule
\end{tabular}
}
\vspace{-5pt}

\caption{Generation Performance of our proposed method.}
\label{tab:mmcd_different_models}
\end{table}

\begin{table}[!t]
\centering
\small

% \scalebox{1}{% 放大到110%，可根据需要调整系数（如1.2/1.3）
% \vspace{-5pt}
\begin{tabular}{lccccc} % 新增k value列，列格式从lcccc改为lccccc
\toprule
\textbf{Strategy} & \textbf{k value} & \multicolumn{4}{c}{\textbf{with rich comments}} \\
\cmidrule(lr){3-6} % 调整分隔线范围，覆盖4个指标列（原2-5列改为3-6列）
 &  & F1 & Rec. & Prec. & Acc. \\
\midrule
\textbf{Top-$k$} 
  & 20 & 70.23 & 70.85 & 69.62 & 69.96 \\
  & 25 & 69.78 & 68.55 & 71.06 & 70.32 \\
  & \textbf{30} & \textbf{71.64} & \textbf{72.08} & \textbf{71.20} & \textbf{71.47} \\
  & 35 & 69.33 & 69.08 & 69.57 & 69.43 \\
\midrule
\textbf{Random-$k$}  & 20 & 68.24 & 66.61 & 69.94 & 68.99 \\
  & 25 & 68.73 & 68.55 & 68.92 & 68.82 \\
 & 30 & 70.40 & 70.23 & 70.85 & 69.98 \\
  & 35 & 71.16 & 70.49 & 72.79 & 69.59 \\
\midrule
\textbf{Full set} & - & 69.81 & 70.67 & 68.97 & 69.43 \\
\bottomrule
\end{tabular}
% }
% \vspace{-5pt}
\caption{Results of different comment sampling strategy.}
\label{tab:mmcd_comments_numbers}
\end{table}

\subsection{Ablation Study}
To evaluate the contribution of each component, we conduct ablation experiments, and the results are shown in Table \ref{tab:mmcd_ablation_study}. The results show  that all agents in AuDisAgent play critical roles in MCD. Removing any single agent leads to a noticeable performance degradation, which highlights the effectiveness of each component. Moreover, the synergy among these agents significantly enhances the framework’s robustness in detecting controversial content.

\subsection{Generalization Experiments}
To evaluate the generalization ability of AuDisAgent, we integrate it with several mainstream LLMs, including GLM-4-9B \cite{glm2024chatglm}, Qwen2.5-7B \cite{qwen2025qwen25technicalreport}, deepseek \cite{deepseekai2025deepseekr1incentivizingreasoningcapability}, and GPT-4o \cite{openai2024gpt4osystem}.
We compare AuDisAgent with AgentMCD, a leading zero-shot multi-agent framework in this field. As shown in Table \ref{tab:mmcd_different_models}, AuDisAgent achieves performance gains across all base models, demonstrating strong adaptability to different LLMs. Notably, the largest improvement is observed on GLM-4-9B, where the F1-score increased by 3.73\%, Recall by 4.15\%, Precision by 3.22\%, and Accuracy by 3.54\%. Overall, AuDisAgent outperforms the current SOTA framework AgentMCD in the majority of scenarios.

\subsection{Comment Sampling Strategy}

The construction of the reference database in the Viewing Panel Agent relies on the analysis of video and comment content. Therefore, sampling strategy is important.
To study this effect, we conduct experiments with different sampling settings.
Given that the MMCD dataset contains about 40 valid comments per video on average, we test four sampling sizes ($k=20,25,30,35$) under two sampling strategies: Top‑$k$ (selecting the most liked comments) and Random‑$k$ (randomly sampling), and we treat the full comment set as the baseline. The results are shown in Table \ref{tab:mmcd_comments_numbers}.
The results indicate that the ``Top‑30" strategy achieves the best overall performance. This suggests that increasing the number of comments does not necessarily improve performance: insufficient sampling may miss key opinions, while excessive sampling tends to introduce noise.

\subsection{Audience Dissemination Simulation}
To further analyze the Viewing Panel Agent, we design two variants of AuDisAgent:
1) \textbf{No Discussion:} Directly construct the initial opinions generated by each audience into the final opinion pool without any discussion.
2) \textbf{Generic Roles:} Discard the diverse audience persona design and uniformly use three generic roles: supporters, opponents, and neutrals.
As shown in Table \ref{tab:mmcd_detailed_discuss}, both variants perform worse than the original design. The ``No Discussion" variation lacks opinion refinement nad thus fails to construct a high-quality opinion pool. The ``Generic Roles" variant oversimplifies audience diversity, producing generic viewpoints that cannot capture core controversies. These results further velidate the effectiveness of the Viewing Panel Agent.

\begin{figure}[!t]
 \small
  \centering
  \includegraphics[width=0.75\linewidth]{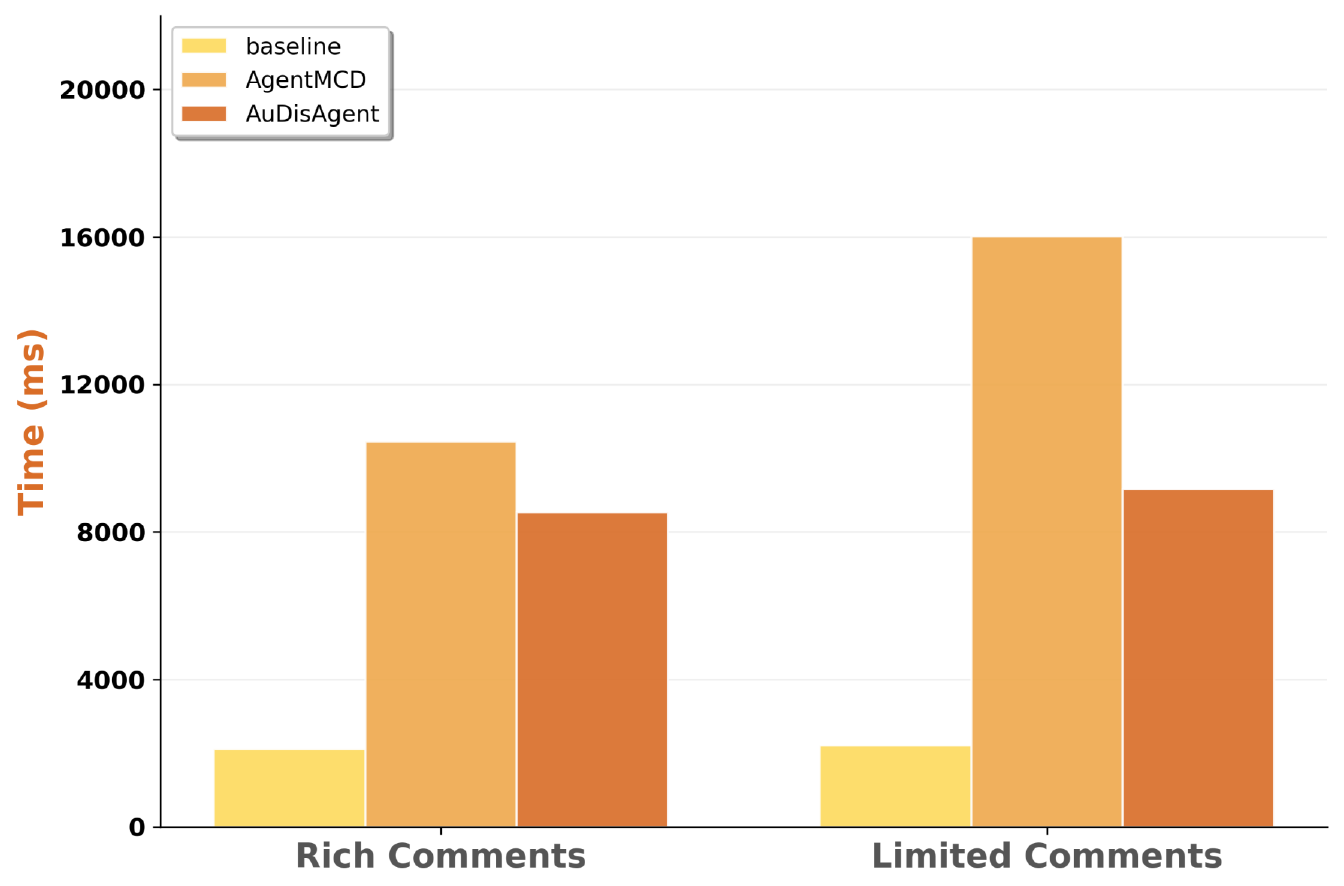}  
  % \vspace{-5pt}
  \caption{Runtime Comparison under Different Comment Settings.}
    % \vspace{-5pt}
  \label{fig:time}
\end{figure}

\begin{figure}[!t]
 \small
  \centering
  \includegraphics[width=\linewidth]{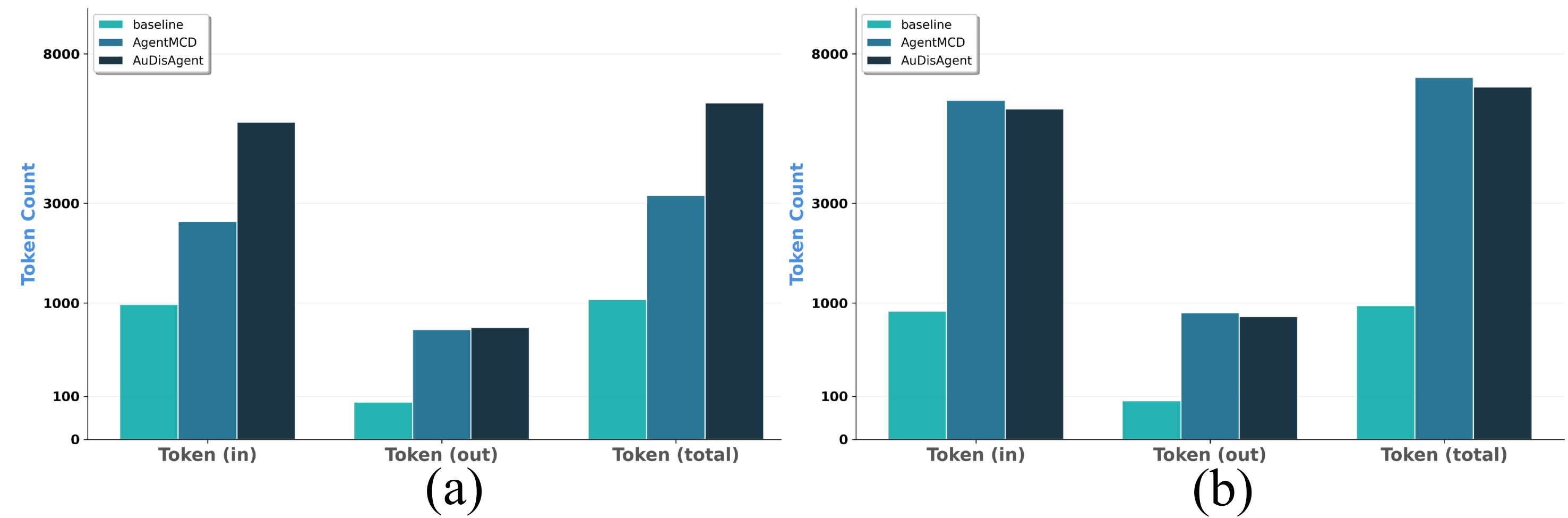}  
  \caption{Token Cost under Different Comment Settings: (a) Rich-comment scenario; (b) Limited-comment scenario.}
  % \vspace{-15pt}
  \label{fig:token}
\end{figure}

\begin{table}[!t]

\centering
\small % 在这里加入 \small 让字体整体小一号
\renewcommand{\arraystretch}{1.05} 
\begin{tabular}{lcccc}
\toprule
\textbf{Method} & F1 & Rec. & Prec. & Acc. \\
\midrule
\multicolumn{5}{c}{\textbf{with rich comments}} \\
\midrule
\textbf{AuDisAgent} & \textbf{71.64} & \textbf{72.08} & \textbf{71.20} & \textbf{71.47} \\
No Discussion       & 70.34 & 72.08 & 68.69 & 69.61 \\
Generic Roles       & 69.72 & 71.38 & 68.13 & 68.99 \\
\midrule
\multicolumn{5}{c}{\textbf{with limited comments}} \\
\midrule
\textbf{AuDisAgent} & \textbf{68.29} & \textbf{69.75} & \textbf{66.89} & \textbf{67.56} \\
No Discussion       & 66.27 & 68.55 & 64.13 & 65.11 \\
Generic Roles       & 65.24 & 67.14 & 63.44 & 64.22 \\
\bottomrule
\end{tabular}
\caption{Performance comparison among different methods.}
\label{tab:mmcd_detailed_discuss}
\end{table}

\subsection{Comment Bootstrapping Strategy}
To investigate the impact of different hyperparameters and strategies in the Comment Bootstrapping Strategy, we conduct sufficient experiments, and the results are shown in Table \ref{tab:mmcd_limited_comments_hyperparams}.

% 为了全面评估Comment Bootstrapping Strategy 中各项超参数和strategy对 AuDisAgent 性能的影响，我们进行了一系列实验，结果汇总于表 6

\noindent \textbf{Reference Database Scale:}
Under limited-comment conditions, we evaluate the impact of different reference database scales by randomly sampling 10\% to 70\% of the full database. The results show a clear positive correlation between database scale and performance. As the database grows, the retrieval module is more likely to identify similar historical samples, resulting in higher-quality migrated comments.

\noindent  \textbf{Encoding Weight:} 
We study the impact of encoding strategies on AuDisAgent during similar-sample matching process by varying the fusion weight between title and keywords. The results show that relying solely on title (title : keywords = 1:0) achieves the best performance, because title often contains more specific information, while keywords are frequently generic and provide limited discriminative information.

\begin{figure*}[!t]
 \small
  \centering
  \includegraphics[width=\linewidth]{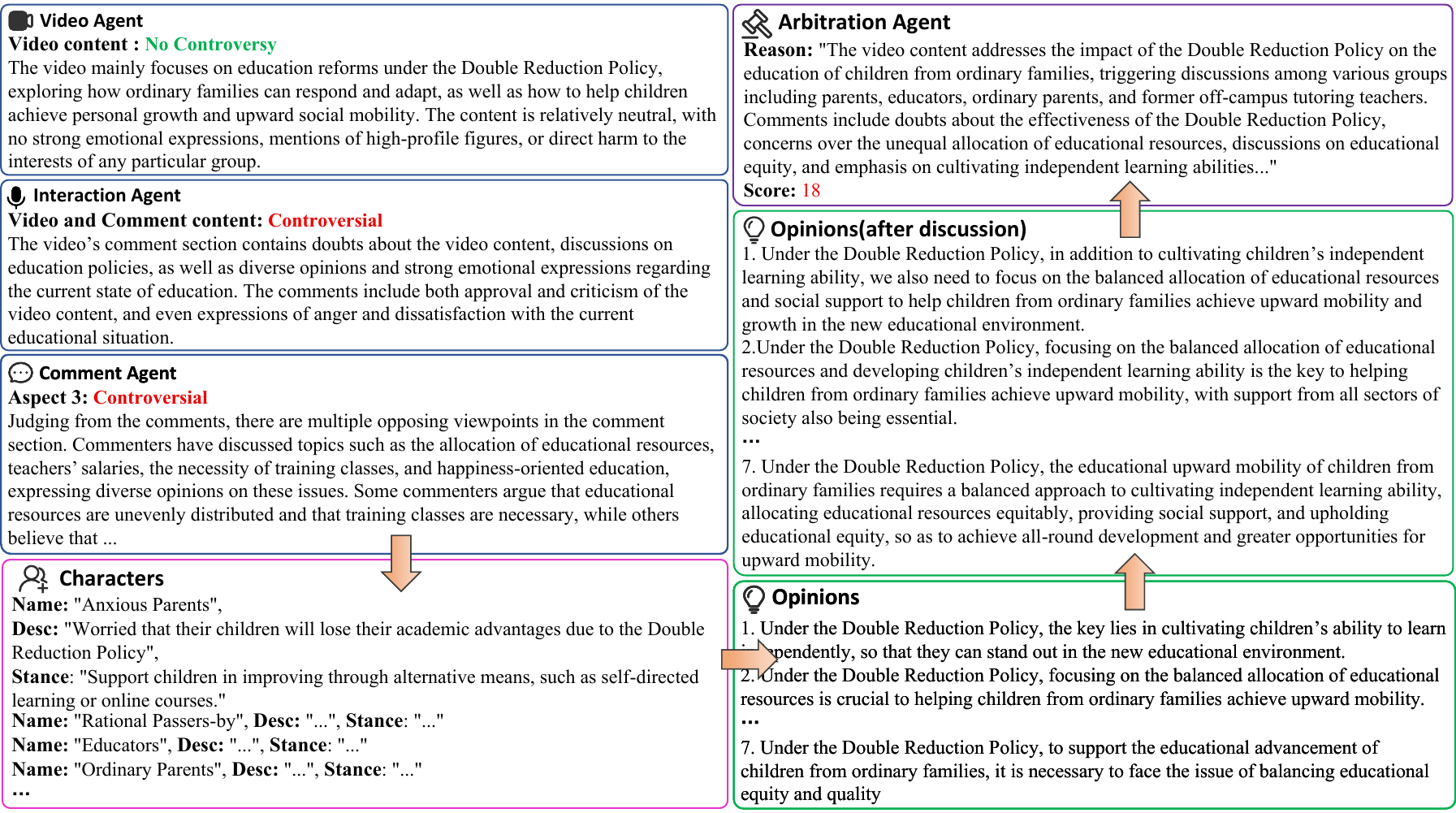}  
  \caption{The complete process of a correctly predicted instance}
  % \vspace{-5pt}
  \label{fig:casestudy}
\end{figure*}

\noindent  \textbf{Comment Migration:} 
We evaluate different comment migration strategies in the Comment Bootstrapping Strategy, extracting the most liked comments from the most similar samples. To ensure fairness, we keep the total number of selected comments fixed at 30. Results show that selecting the top-10 most-liked comments from top-3 similar samples (Top3–10) achieves the best performance. 
This is because insufficient samples limit comment diversity and increase the risk of overfitting to single-video contexts, while excessive samples introduce noise or too few comments per sample, hindering the simulation of complete discussions.

\noindent \textbf{Embedding Model:} 
% To investigate the impact of different embedding1 models on the performance of AuDisAgent,
We compare several pre-trained embedding models, including all-MiniLM-L6-v2 \cite{wang2020minilmdeepselfattentiondistillation}, e5-large-v2 \cite{wang2024textembeddingsweaklysupervisedcontrastive} bge-large-zh-1.5 \cite{xiao2024cpack}, and GPT-3 \cite{GPT-3} (used as the text encoder in mPLUG-Video). Results show that bge-large-zh-1.5 achieves the best performance, because it is optimized for Chinese semantic scenarios. Therefore, when extending the framework to other languages, we recommend using language-specific encoders as the base model.

\begin{table}[!t]
\centering

% === 核心技巧：去除局部横线自带的上下留白，依靠 arraystretch 保证所有行绝对等高 ===
\setlength{\aboverulesep}{0pt}
\setlength{\belowrulesep}{0pt}
\renewcommand{\arraystretch}{1.25} 

\resizebox{\columnwidth}{!}{%
\begin{tabular}{llcccc}
\toprule
\textbf{Parameter Category} & \textbf{Setting} & \textbf{F1} & \textbf{Rec.} & \textbf{Prec.} & \textbf{Acc.} \\
\midrule
% 第 1-5 行
\multirow{5}{*}{\begin{tabular}[c]{@{}l@{}}\textbf{Reference Database} \\ \textbf{Scale}\end{tabular}} 
 & 10\% & 61.75 & 63.60 & 63.05 & 63.16 \\
 & 30\% & 63.85 & 62.72 & 65.02 & 64.49 \\
 & 50\% & 64.09 & 63.07 & 65.15 & 64.66 \\
 & 70\% & 65.02 & 65.19 & 64.85 & 64.93 \\
 & \textbf{100\%} & \textbf{68.29} & \textbf{69.75} & \textbf{66.89} & \textbf{67.56} \\
\cmidrule(lr){1-6} 
% 第 6-10 行
\multirow{5}{*}{\begin{tabular}[c]{@{}l@{}} \textbf{Encoding Weight} \\ \textbf{(title : keywords)} \end{tabular}} 
 & 0:1 & 67.36 & 68.37 & 66.38 & 66.87 \\
 & 0.25:0.75 & 67.30 & 68.37 & 66.27 & 66.78 \\
 & 0.5:0.5 & 66.49 & 67.14 & 65.86 & 66.17 \\
 & 0.75:0.25 & 67.25 & 68.02 & 66.49 & 66.87 \\
 & \textbf{1:0} & \textbf{68.29} & \textbf{69.75} & \textbf{66.89} & \textbf{67.56} \\
\cmidrule(lr){1-6} 
% 第 11-14 行
\multirow{5}{*}{\begin{tabular}[c]{@{}l@{}} \textbf{Comment Migration} \\ \textbf{(sample - comments)} \end{tabular}}

 & top1-30 & 64.69 & 63.60 & 65.81 & 65.28 \\
 & top2-15 & 65.43 & 65.55 & 65.32 & 65.37 \\
 & \textbf{top3-10} & \textbf{68.29} & \textbf{69.75} & \textbf{66.89} & \textbf{67.56} \\
 & top5-6 & 66.08 & 65.57 & 66.61 & 65.81 \\
 & top6-5 & 66.02 & 65.62 & 66.43 & 65.81 \\
\cmidrule(lr){1-6} 
% 第 15-17 行 (新增：Embedding Model)
\multirow{4}{*}{\textbf{Embedding Model}} 
 & all-MiniLM-L6 & 63.95 & 65.61 & 62.37 & 64.84 \\
 & e5-large & 63.16 & 64.93 & 61.48 & 64.13 \\
 & GPT-3 & 63.58 & 66.47 & 60.95 & 65.11 \\
 & \textbf{bge-large-zh} & \textbf{68.29} & \textbf{69.75} & \textbf{66.89} & \textbf{67.56} \\
\cmidrule(lr){1-6} 
% 第 18-20 行
\multirow{3}{*}{\textbf{Overall}} 
 & Baseline & 64.39 & 64.40 & 64.42 & 64.40 \\
 & AgentMCD & 66.29 & 66.45 & 66.78 & 66.46 \\
 & \textbf{AuDisAgent} & \textbf{68.29} & \textbf{69.75} & \textbf{66.89} & \textbf{67.56} \\
\bottomrule
\end{tabular}%
}
\caption{Performance comparison of different hyperparameters under the limited comments condition.}
\label{tab:mmcd_limited_comments_hyperparams}
\end{table}

\subsection{Computational Overhead}

To evaluate the computational efficiency, we report the average results over 100 randomly sampled instances using a single NVIDIA RTX 4090D GPU. The results are shown in Figs. \ref{fig:time} and \ref{fig:token}. Under the rich comment scenario, although AuDisAgent consumes more tokens than AgentMCD, it acheives faster runtime. This is because AgentMCD adopts more time-consuming scoring strategies at each stage and applies the same workflow to both simple and ambiguous samples. Under the limited comment scenario, AuDisAgent is also faster than AgentMCD. This efficiency advantage mainly stems from AuDisAgent's comment bootstrapping strategy, which avoids the heavy overhead of AgentMCD's agent-based comment simulation.

\begin{table}[t]
  \centering
 
  \resizebox{.95\columnwidth}{!}{%
  \begin{tabular}{lcccc}
    \toprule
    \multirow{2}{*}{\textbf{Metric}} & \multicolumn{2}{c}{\textbf{Baseline}} & \multicolumn{2}{c}{\textbf{AudisAgent}} \\
    \cmidrule(lr){2-3} \cmidrule(lr){4-5}
    & \textbf{RC} & \textbf{LC} & \textbf{RC} & \textbf{LC} \\
    \midrule
    Faithfulness         & 7.28 & 6.97 & 7.93 & 7.61 \\
    Inference Coherence  & 7.13 & 6.92 & 7.71 & 7.38 \\
    Inference Depth      & 5.84 & 5.75 & 6.93 & 6.74 \\
    Judgment Rationality & 6.69 & 6.46 & 7.37 & 7.06 \\
    Expression Clarity   & 8.01 & 7.89 & 8.23 & 8.05 \\
    \bottomrule
    % \multicolumn{5}{l}{\small $^{\dagger}$ RC: Rcih Comment; LC: limited Comment.} \\
  \end{tabular}%
  }
   \caption{Interpretability Evaluation}
  \label{tab:interpretability_eval}
\end{table}
\subsection{Case Study}
To illustrate the reasoning chain of AuDisAgent, we present a representative case in Figure \ref{fig:casestudy}. AuDisAgent first performs a preliminary assessment through three screening agents. When consensus cannot be reached, the Viewing Panel Agent is activated to further analyze the controversial content and facilitate the final classification. Each step in the reasoning process is rigorous and logical. Through this structured reasoning process, the model ultimately assigns a reasonable controversy score to the example and produces the correct judgment, demonstrating the interpretability of AuDisAgent’s decision-making.

\subsection{Interpretability Evaluation}
Following recent works \cite{Chiang_Gonzalez_Li_Li_Lin_Sheng_Stoica_Wu_Xing_Zhang_et},  we evaluate the interpretability of AuDisAgent's reasoning chains. Specifically, we randomly sample 100 chains and use Qwen3.5-flash \cite{qwen3.5_tech_report} to score five dimensions: 1) \textit{Faithfulness:} whether the output is faithful to the input and free of hallucinations; 2) \textit{Inference Coherence:} the logical rationality of the model's reasoning chain; 3) \textit{Inference Depth:} whether the output can address the core controversial points 4) \textit{Judgment Rationality:} the reasonableness of the model's decision; 5) \textit{Expression Clarity:} the linguistic quality and fluency of the generated text. Each of them are scored on a 0-10 scale, and the results are reported in Table \ref{tab:interpretability_eval}. 

The results show that AuDisAgent outperforms all baselines under both Rich Comment (RC) and Limited Comment (LC) settings, with notable gains in Inference Depth and Judgment Rationality, demonstrating its high interpretability across all dimensions.

\section{Conclusion}
In this paper, we introduce AuDisAgent, a training-free multi-agent framework that overcomes the limitations of static feature modeling in MCD. To address challenges such as dynamic public opinion evolution and the cold-start problem for newly released videos, we design a dynamic audience dissemination simulation combined with a Comment Bootstrapping Strategy. Comprehensive experiments on the public MMCD dataset demonstrate the effectiveness of our framework in both rich- and limited-comment scenarios. In future work, we plan to extend AuDisAgent to other reasoning-based multimodal tasks, exploring its broader applicability in content understanding.

\section{Limitations}
Although our proposed achieved satisfactory performance, there are several ways to further improve this work:

1) The original design of AudisAgent is mainly targeted at controversy detection for short‑form video content such as TikTok and YouTube Shorts. To extend this method to the long‑form video domain, more sophisticated visual‑semantic alignment modules and keyframe extraction techniques may be required to accurately locate controversial segments in long sequential timelines.

2) Limited by the inherent constraints of the dataset, AudisAgent performs retrieval using title information from sample metadata. While it has already achieved decent performance, with more comprehensive datasets available in the future, AudisAgent could adopt additional information or retrieval strategies to further boost detection accuracy.

3) Although AudisAgent integrates multiple strategies to deeply mine core controversies and enhance model understanding, its final detection performance is still constrained by the capability bottlenecks of the underlying foundation models. In the future, adopting models with stronger capabilities in capturing socio-cultural differences or more superior performance is expected to further improve the accuracy of detection.

\section{Ethics Statement}

This study aims to combat online controversies and public opinion conflicts through training-free multimodal detection methods, contributing to building a safer online space for social video platforms. The types of controversial content focused on in this study are core issues that are critical for content governance and risk management on short-video platforms such as TikTok and YouTube Shorts. Our work concentrates on detecting potential risks of public opinion conflict and polarized views that can rapidly escalate into reputational damage or social conflicts.
However, we also recognize that malicious actors may use reverse engineering techniques to craft video content or manipulate comments in order to evade detection by AI systems like AuDisAgent or cause them to make misjudgments. We firmly condemn such behaviors and emphasize that this study is solely for the purposes of scientific research and controversial content prevention. The relevant framework and supporting resources are strictly prohibited from being used for commercial profit or malicious abuse.
To ensure the responsible development and evaluation of the framework, we have implemented a number of protective measures:
1) All experiments use publicly available research datasets and fully comply with the usage agreements of each dataset;
2) No personal user data has been independently collected for this study; 
We believe that the benefits of improving multimodal controversy detection capabilities far outweigh the potential risks—especially at a time when short-video platforms can serve as breeding grounds for online controversies. It should also be noted that the views, stances, and diverse opinions contained in the video and comment samples do not represent the positions of the study's authors. The framework we designed is intended to assist rather than replace human content moderation work, maintaining a healthy online community environment through collaborative efforts.

\bibliography{custom}

\appendix

\end{document}